\documentclass[letterpaper]{article} % DO NOT CHANGE THIS
\usepackage{aaai2027}
\usepackage[hyphens]{url}  % DO NOT CHANGE THIS
\usepackage{graphicx} % DO NOT CHANGE THIS
\usepackage{natbib}  % DO NOT CHANGE THIS AND DO NOT ADD ANY OPTIONS TO IT
\usepackage{caption} % DO NOT CHANGE THIS AND DO NOT ADD ANY OPTIONS TO IT
\usepackage{xcolor}
\usepackage{colortbl}
\usepackage{pifont}
\newcommand{\cmark}{\textcolor{green!50!black}{\ding{51}}} 
\newcommand{\xmark}{\textcolor{red!80!black}{\ding{55}}}
\newcommand{\pmark}{\textcolor{black!55}{--}}

\usepackage{amsmath}
\usepackage{amsfonts}
\usepackage{booktabs}
\usepackage{multirow}
\usepackage{array}

\newcolumntype{C}[1]{>{\centering\arraybackslash}p{#1}}

\nocopyright

\title{Are LLMs Ready for Scientific Discovery? A Capability-Oriented Benchmark for AI Scientists}
\author{
    Chuhan Shi\textsuperscript{\rm 1},
    Xiaoquan Ren\textsuperscript{\rm 1},
    Sicheng Song\textsuperscript{\rm 2},
    Haobo Li\textsuperscript{\rm 3},\\
    Rui Sheng\textsuperscript{\rm 3}\corresponding,
    Yushi Sun\textsuperscript{\rm 3}\corresponding
}

\affiliations{
    \textsuperscript{\rm 1}Southeast University\\
    \textsuperscript{\rm 2}East China Normal University\\
    \textsuperscript{\rm 3}Hong Kong University of Science and Technology\\
    chuhanshi@seu.edu.cn,
    renxiaoquan987@gmail.com,
    scsong@dase.ecnu.edu.cn,\\
    rshengac@connect.ust.hk,
    ysunbp@connect.ust.hk
}

\begin{document}

\maketitle

\begin{abstract}
Existing benchmarks for scientific data analysis evaluate LLMs primarily on code execution or workflow completion, overlooking that scientific analysis serves to support distinct types of scientific claims: hypothesis exploration, statistical inference, mechanistic explanation, each with different assumptions and validity criteria. We introduce \textbf{SDABench}, a benchmark that reorganizes evaluation around six capabilities (descriptive, exploratory, inferential, predictive, causal, and mechanistic) across five domains (Biology, Chemistry, Environment, Geography, Physics). SDABench comprises 527 real-data instances (SDA-Real) and 6{,}000 synthetic instances (SDA-Synth), each in both multiple-choice and open-ended formats, constructed through an automated pipeline. Evaluating 15 representative LLMs, we find that models handle descriptive analysis well but degrade sharply on tasks requiring assumption selection, latent-process modeling, or mechanistic reasoning. SDABench further provides a five-stage error analysis framework that locates where LLMs fail: more advanced models more reliably identify the relevant scope and variables, but still struggle to select appropriate analytical procedures, model variable relationships, and draw valid conclusions.
\end{abstract}

\section{Introduction}

Large language models (LLMs) are increasingly transforming scientific discovery by assisting researchers with tasks ranging from protein structure prediction~\cite{alphafold} to automated hypothesis generation~\cite{hypothesismaterials,airesearcher}. At the core of this discovery pipeline is scientific data analysis~\cite{shi2026survey}, the phase that converts raw experimental, simulated, or observational data into validated scientific knowledge~\cite{position,discoverybench}. Ensuring the reliability of AI-driven science therefore requires systematically evaluating how effectively LLMs analyze and reason with scientific data.

However, existing benchmarks such as DataSciBench~\cite{datascibench}, KramaBench~\cite{kramabench}, and ScienceAgentBench~\cite{scienceagentbench} primarily evaluate scientific data analysis from a computational perspective. They assess whether a system can generate and execute analysis code or pipelines, such as selecting features and building models, scoring the results against ground-truth answers or task-specific success criteria. While these benchmarks have expanded the realism and scope of data analysis evaluation, they carry two critical limitations. First, they can mistake successful execution for scientific validity: computing the requested result does not mean that the result supports the intended scientific claim. Second, they hide specific reasoning failures behind aggregate scores: a single workflow score cannot reveal whether a model fails at basic description or at complex causal inference. 
Scientific data analysis ultimately serves not merely to build models or maximize accuracy but to support the scientific claims drawn from data. These claims take different forms, including description, hypothesis exploration, statistical inference, prediction, causal estimation, and mechanistic explanation~\cite{leek2015what}. They require different assumptions, evidence boundaries, and validity criteria, even when they rely on similar analytical procedures. Therefore, the evaluation for scientific data analysis should be organized around the claim types a model can reason about, rather than only the operations it can perform. This motivates a capability-oriented framework that evaluates LLMs by the types of scientific questions they can reason about from data, so as to characterize their strengths and weaknesses across distinct modes of scientific reasoning.

\begin{table*}[t]
\centering
{\small
\setlength{\tabcolsep}{3.5pt}
\renewcommand{\arraystretch}{1.05}
\begin{tabular*}{\textwidth}{@{\extracolsep{\fill}} l c | c c c c c c c @{}}
\toprule
\textbf{Benchmarks} & \textbf{Ours} & \textbf{QRData} & \textbf{LLM-SR} & \textbf{DSBench} & \textbf{DA-Code} & \textbf{Krama} & \textbf{BLADE} & \textbf{SciAgent} \\
\midrule
\rowcolor{gray!15}
\multicolumn{9}{@{}l}{\textbf{Ability Coverage}}\\
Descriptive reasoning  & \cmark & \cmark & \xmark & \cmark & \cmark & \cmark & \cmark & \cmark \\
Exploratory reasoning  & \cmark & \xmark & \xmark & \xmark & \cmark & \cmark & \xmark & \xmark \\
Inferential reasoning  & \cmark & \cmark & \xmark & \xmark & \cmark & \xmark & \cmark & \cmark \\
Predictive reasoning   & \cmark & \xmark & \xmark & \cmark & \cmark & \cmark & \xmark & \cmark \\
Causal reasoning       & \cmark & \cmark & \xmark & \xmark & \xmark & \xmark & \xmark & \xmark \\
Mechanistic reasoning  & \cmark & \xmark & \cmark & \xmark & \xmark & \xmark & \xmark & \xmark \\
\midrule
\rowcolor{gray!15}
\multicolumn{9}{@{}l}{\textbf{Benchmark Characteristics}}\\
Format          & MCQ+OEQ & MCQ+OEQ & OEQ & Exec & Exec & Exec & MCQ+OEQ & Exec \\
Scale (\#inst.)  & 6{,}527 & 411 & 239 & 540 & 500 & 104 & 724 & 102 \\
Data source     & Real+Synth & Real & Real+Synth & Real & Real & Real & Real & Real \\
Evaluation      & Auto+LLM & Auto & Auto+LLM & Auto+LLM & Auto & Auto+LLM & Auto+LLM & Auto+LLM \\
Error analysis  & \cmark & \cmark & \xmark & \pmark & \pmark & \pmark & \pmark & \cmark \\
\bottomrule
\end{tabular*}
}
\caption{Comparison with representative data analysis benchmarks. Format abbreviations: MCQ = multiple-choice, OEQ = open-ended, Exec = execution-based. ``\pmark'' indicates partial satisfaction of structured error categorization.}
\label{tab:bench-compare}
\end{table*}

To address this problem, we adopt the six-question framework of Leek and Peng~\cite{leek2015what} and reinterpret its data analysis question types as six scientific discovery capabilities: descriptive, exploratory, inferential, predictive, causal, and mechanistic. 
Based on this taxonomy, we construct \textbf{SDABench} across five core scientific domains: Biology, Chemistry, Environment, Geography, and Physics. It consists of two complementary components: SDA-Real, a curated collection of real-world scientific datasets, and SDA-Synth, which contains synthetic datasets generated through the construction of semantic graphs. Utilizing a three-stage construction pipeline of LLM-based template construction, programmatic ground-truth derivation, and option generation, SDABench includes 527 instances in SDA-Real and 6,000 instances in SDA-Synth. Each instance is provided in both multiple-choice question (MCQ) and open-ended question (OEQ) formats. 
Furthermore, we evaluate 15 representative LLMs and analyze how model errors vary across task types, domains, and model scales. Two findings stand out from this evaluation. First, providing explicit candidate options can degrade performance on certain task types by inducing superficial pattern matching, calling into question the widespread reliance on MCQ-based evaluation. Second, although more advanced models more often identify the task objectives and relevant variables, they continue to encounter difficulties in selecting appropriate analyses, accurately capturing relationships among variables, and ensuring that conclusions are supported by the evidence. Aggregate workflow scores obscure these shortcomings, which become apparent only when evaluations differentiate scientific capabilities and analyze the reasons behind each incorrect response. The main contributions of this work are:

\begin{itemize}
    \item We introduce \textbf{SDABench}, a benchmark for scientific data analysis encompassing five scientific domains and six scientific discovery capabilities. SDABench includes 527 real-data instances and 6,000 synthetic instances, each provided in both MCQ and OEQ formats.
    \item We develop an automated benchmark construction pipeline that converts scientific resources into verified analysis instances through LLM-based template construction, programmatic ground-truth derivation, and option generation.
    \item We propose a five-stage error analysis framework, which classifies failures along the execution pipeline of scientific data analysis (Scope, Variable, Function, Relationship, Conclusion). This taxonomy reveals that as models improve, early grounding errors decrease while later-stage errors persist, making Function, Relationship, and Conclusion Errors the dominant failure modes for stronger models, a diagnostic signal invisible to aggregate accuracy scores.
\end{itemize}

\section{Related Work}

\paragraph{Benchmarks for Data Analysis.}
Existing data analysis benchmarks (Table~\ref{tab:bench-compare}) span three settings: execution-oriented code generation (DS-1000~\cite{ds1000}; DSCodeBench~\cite{dscodebench}), multi-step data-science workflows (DA-Code~\cite{dacode}; DSBench~\cite{dsbench}; DataSciBench~\cite{datascibench}), and data-grounded scientific reasoning (QRData~\cite{qrdata}; KramaBench~\cite{kramabench}; BLADE~\cite{blade}; ScienceAgentBench~\cite{scienceagentbench}; LLM-SRBench~\cite{llmsrbench}). Most are organized around execution success, pipeline completion, or domain-specific tasks, which blurs failures in basic data manipulation with failures in deeper scientific reasoning and makes it difficult to diagnose which analytical capability a model actually lacks.

\paragraph{LLMs for Data Analysis.}
Data-analysis assistants have transitioned from basic exploratory tools~\cite{insightpilot, sheetcopilot, datacopilot} to autonomous agents capable of complex planning and verification~\cite{dsagent, datainterpreter, autoprep, dsstar, deepanalyze}. However, successful workflow completion does not guarantee that an agent can validate statistical assumptions, handle uncertainty, or model scientific mechanisms. This shifts the evaluation bottleneck from execution success to distinct analytical capabilities, requiring a diagnostic framework to independently assess these reasoning dimensions.

\section{Benchmark}
\label{sec:benchmark}

\subsection{Overview}
\label{sec:overview}

We introduce SDABench, a benchmark designed to systematically evaluate the scientific data analysis capabilities of LLMs. SDABench covers five scientific domains (Biology, Chemistry, Environment, Geography, and Physics) and six scientific discovery capabilities, which we operationalize as task types: \emph{descriptive} tasks summarize measurements to characterize a phenomenon; \emph{exploratory} tasks examine measurements to discover patterns that may suggest testable hypotheses; \emph{inferential} tasks use sampled measurements to evaluate a scientific claim about a broader phenomenon with quantified uncertainty; \emph{predictive} tasks use measured variables to estimate unknown observations or future states; \emph{causal} tasks estimate how a defined intervention, exposure, or perturbation changes an outcome; and \emph{mechanistic} tasks use data and domain theory to explain the process that produces an observed scientific outcome.

Each benchmark instance is formalized as $z=(d,t,D,Q,g)$, where $d$ is the scientific domain, $t$ is the task type, $D$ denotes the referenced dataset and its metadata used to answer the question, $Q$ is the question, and $g$ is the ground-truth answer. For every instance, we release two formats: multiple-choice questions $Q_{\mathrm{MCQ}}(z)$, which support scalable automatic evaluation through plausible distractors, and open-ended questions $Q_{\mathrm{OEQ}}(z)$, which ask models to produce the answer directly and are better suited for open-form reasoning and analysis execution.

\subsection{Data Curation}
\label{sec:data-curation}

\begin{figure}[t]
  \centering
  \includegraphics[width=\linewidth]{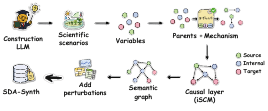}
  \caption{The SDA-Synth construction pipeline.}
  \label{fig:sda-synth-pipeline}
\end{figure}

We curate SDABench by screening public scientific datasets and benchmarks using three criteria. First, they should target quantitative scientific analyses with clear objectives. Second, they are required to provide structured data with clear metadata. Third, their answers should be programmatically derivable from the underlying data. Based on these criteria, we select KramaBench~\cite{kramabench}, BLADE~\cite{blade}, LLM-SRBench~\cite{llmsrbench}, QRData~\cite{qrdata}, DiscoveryBench~\cite{discoverybench}, SciTS~\cite{scits}, and ScienceAgentBench~\cite{scienceagentbench} as seed resources from 24 candidate scientific datasets and benchmarks. We use these resources to construct two dataset components: \textbf{SDA-Real}, a re-curated collection of real scientific datasets, and \textbf{SDA-Synth}, a generated collection of synthetic scientific datasets. Figure~\ref{fig:sda-synth-pipeline} summarizes the construction pipeline.

\paragraph{SDA-Real.}
SDA-Real is constructed from eligible real datasets drawn from the selected seed resources. We retain a dataset only when its schema and metadata support at least one of the six task types. Rather than reusing the original questions or answers, we generate new questions and derive their answers using the task taxonomy and construction pipeline described in Section~\ref{sec:construction}.

\paragraph{SDA-Synth.}
Real scientific datasets are typically study-specific, making it difficult to evaluate all six task types on a single dataset, and using different datasets per task type would confound task difficulty with dataset artifacts. Following prior work~\cite{discoverybench,llmsrbench,iscm}, we construct synthetic datasets shared across task types, as illustrated in Figure~\ref{fig:sda-synth-pipeline}.
\textbf{1) Scientific problem selection.} We prompt the construction LLM (GPT-4o) to propose candidate scientific scenarios across our five domains (e.g., plant stress physiology, polymerization kinetics), each specifying measurable variables, admissible units, and ranges.
\textbf{2) Semantic graph construction.} For each scenario, we build a semantic DAG $\mathcal{G}=(V,E)$ whose nodes are source, internal, and target variables, each annotated with metadata (type, unit, range). For each non-source variable, the construction LLM proposes a mechanism conditioned on its parents, implemented as explicit algebraic equations or numerical solver specifications (e.g., ODEs). Variables with causal semantics are generated on a causal sub-DAG $G\subseteq \mathcal{G}$ using internally standardized SCMs (iSCMs)~\cite{iscm}, modeling how interventions propagate through intermediate variables to the target.
\textbf{3) Data generation.} We generate each dataset in topological order over $\mathcal{G}$: source variables are sampled from metadata, and internal/target variables are computed by executing the retained mechanisms under fixed parameters, followed by controlled measurement noise and distractor columns. To support exact ground-truth derivation (Section~\ref{sec:construction}), we store the graph structure, causal sub-DAG, equations, and random seeds as hidden metadata. After execution replay, range checks, and expert review, we obtain 50 validated datasets (10 per domain), each supporting all 6 task types.

\begin{figure*}[t]
  \centering
  \includegraphics[width=\textwidth]{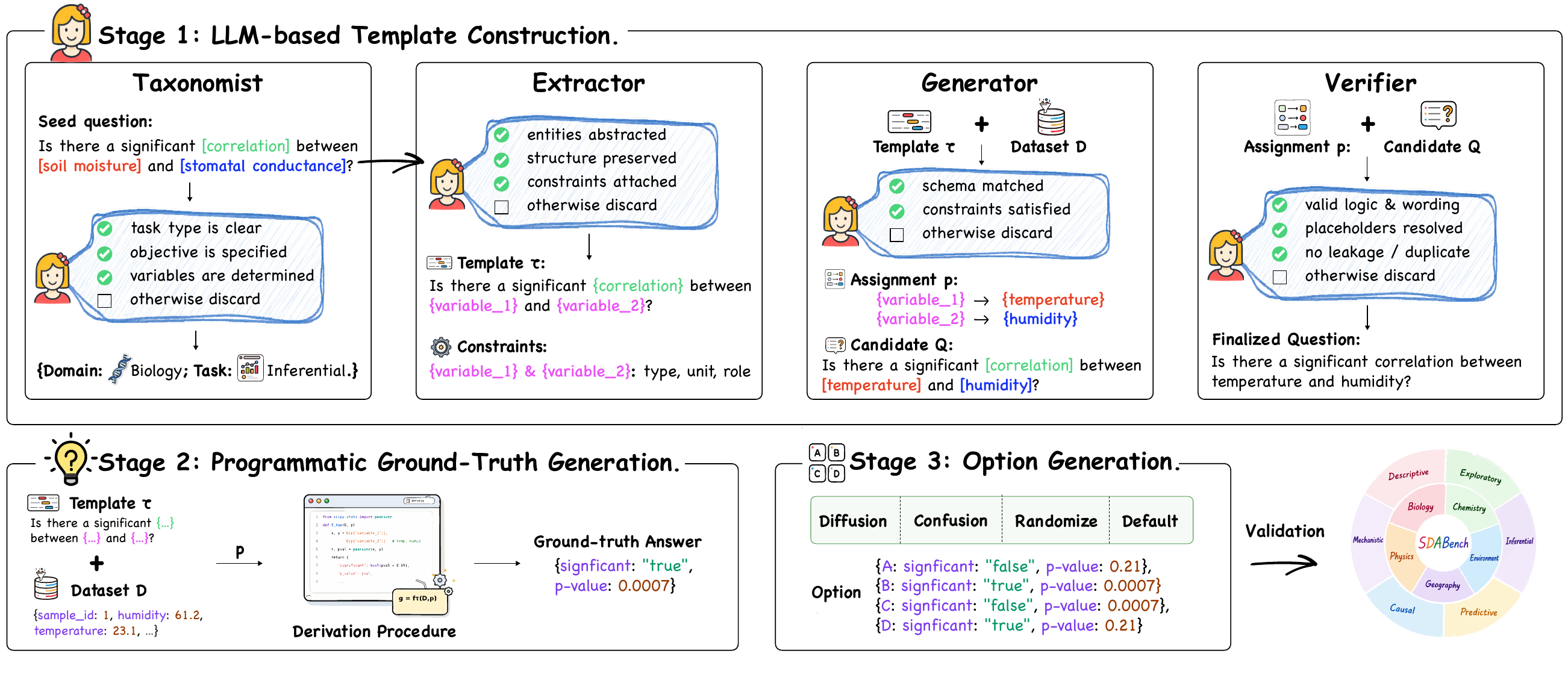}
  \caption{The Question construction pipeline.}
  \label{fig:question-pipeline}
\end{figure*}

\subsection{Question Construction}
\label{sec:construction}

Rather than manually authoring all questions or relying on unconstrained LLM generation, which can produce questions with weak data grounding or unverifiable answers~\cite{toolqa}, we use a construction pipeline inspired by automatic and agentic benchmark construction frameworks~\cite{autobencher,selfevolving}. As shown in Figure~\ref{fig:question-pipeline}, the pipeline builds typed templates from seed analysis patterns, instantiates them on SDA-Synth and SDA-Real datasets with executable answer derivation, and adds perturbation-based distractors for MCQ instances. We retain only instances that pass verification, normalization, deduplication, and targeted human review. 

\paragraph{Stage 1: LLM-based Template Construction.}
We transform seed questions from the selected scientific resources into verified question templates using a four-agent pipeline: a \emph{Taxonomist} assigns each seed question to a domain and one of the six task types and identifies the required data columns; an \emph{Extractor} abstracts dataset-specific entities into a typed template $\tau$ with typed placeholders and validity constraints (e.g., \emph{"Is there a significant correlation between soil moisture and stomatal conductance?''} becomes \emph{"Is there a significant correlation between \{variable\_1\} and \{variable\_2\}?''}); a \emph{Generator} instantiates $\tau$ on a target dataset $D$ by binding each placeholder to a valid column, value, or entity, yielding a candidate question $Q$; and a \emph{Verifier} rejects questions with logical inconsistencies, unresolved placeholders, schema mismatches, answer leakage, or near-duplicates.

\paragraph{Stage 2: Programmatic Ground-Truth Generation.}
To create accurate answers for generated questions, we implement a deterministic derivation procedure $f_{\tau}$ for each template $\tau$. The procedure takes the reference data $D$ and the instantiated parameters $p$ as input and returns the ground-truth answer $g = f_{\tau}(D, p)$. After sampling a valid $p$, we fill the template to form the question and run the corresponding derivation procedure on $D$. Since $p$ contains the exact values used in the question, the procedure can replay the requested analysis and automatically compute the correct answer, even when the answer requires multiple reasoning steps. Before instance generation, all derivation procedures are checked through unit tests and manual review. The OEQ format uses $g$ as the final answer directly. 

\paragraph{Stage 3: Option Generation.}
\label{sec:distractors}
For the MCQ format, we construct plausible but incorrect distractor options through perturbations of variables or intermediate computation steps~\cite{atmosscibench}. The correct option is the answer from the derivation procedure, while incorrect options are produced by four perturbation strategies: \emph{Diffusion} (swapping or reassigning values and roles between variables), \emph{Confusion} (modifying a single local element such as a parameter or operator), \emph{Randomization} (sampling a new task-compliant assignment that yields a different answer), and \emph{Default} (applying simple numeric transformations as a fallback). This design evaluates whether a model follows the intended analysis rather than relying on superficial elimination cues. We randomize option order to reduce positional bias.

  \begin{table*}[t]
    \centering
    {\small
    \setlength{\tabcolsep}{0.45mm}
    \renewcommand{\arraystretch}{1.15}
    \begin{tabular*}{\textwidth}{@{\extracolsep{\fill}}l|*{7}{C{0.32in}}|*{7}{C{0.32in}}@{}}
    \toprule[1.5pt]
    \multirow{2}{*}{\textbf{Model}}
      & \multicolumn{7}{c|}{\textbf{OEQ}}
      & \multicolumn{7}{c}{\textbf{MCQ}} \\
    \cmidrule(lr){2-8}\cmidrule(lr){9-15}
      & {Overall} & D & I & E & P & C & M
      & {Overall} & D & I & E & P & C & M \\
    \midrule
    \multicolumn{15}{c}{\textbf{Open-source LLMs}} \\
    \midrule
    DeepSeek V3.2
      & 51.17 & 88.50 & 59.00 & 41.50 & 31.00 & 46.00 & 41.00
      & 62.21 & 90.00 & 69.00 & 44.00 & 47.75 & 45.75 & 76.75 \\
    DeepSeek R1
      & 42.13 & 85.00 & 46.50 & 34.50 & 23.50 & 40.00 & 23.25
      & 56.38 & 84.25 & 67.00 & 42.25 & 36.50 & 40.50 & 67.75 \\
    Qwen 3.5-397B-A17B
      & 51.58 & 89.00 & 61.50 & 37.00 & 37.00 & 37.00 & \underline{48.00}
      & 62.46 & 92.00 & 73.75 & 44.00 & 35.25 & 52.75 & 77.00 \\
    Qwen 3-235B-Instruct
      & 40.46 & 77.25 & 34.50 & 27.75 & \underline{42.00} & 38.75 & 22.50
      & 52.42 & 85.50 & 59.25 & 37.75 & \underline{49.50} & 36.50 & 46.00 \\
    Kimi K2.5
      & 50.29 & 84.50 & 43.50 & 29.50 & 34.50 & 39.75 & \textbf{\underline{70.00}}
      & 57.04 & 92.25 & 73.00 & 34.50 & 25.75 & 36.25 & \textbf{80.50} \\
    GLM 5
      & 51.67 & 87.75 & 64.75 & 36.50 & \textbf{42.75} & \underline{48.00} & 30.25
      & 60.79 & 91.50 & \textbf{74.50} & 38.75 & 41.00 & 42.00 & 77.00 \\
    Llama 3.3-70B-Instruct
      & 6.96 & 12.00 & 8.00 & 11.75 & 2.75 & 4.50 & 2.75
      & 30.46 & 28.00 & 39.50 & 23.50 & 28.25 & 24.00 & 39.50 \\
    Llama 3.1-8B-Instruct
      & 5.04 & 10.25 & 3.50 & 8.25 & 0.50 & 7.75 & 0.00
      & 25.00 & 21.50 & 17.75 & 21.50 & 24.75 & 30.50 & 34.00 \\
    \midrule
    Average (Open-Source)
      & 37.41 & 66.78 & 40.16 & 28.34 & 26.75 & 32.72 & 29.72
      & 50.84 & 73.13 & 59.22 & 35.78 & 36.09 & 38.53 & 62.31 \\
    \midrule
    \multicolumn{15}{c}{\textbf{Closed-source LLMs}} \\
    \midrule
    GPT-5.4
      & \textbf{\underline{58.33}} & \textbf{\underline{94.75}} & \textbf{\underline{67.25}} & \textbf{\underline{53.75}} & 34.50 & \textbf{\underline{52.75}} & 47.00
      & \textbf{66.58} & \underline{93.50} & \textbf{\underline{78.50}} & 45.50 & \textbf{58.25} & 40.75 & \textbf{\underline{83.00}} \\
    GPT-5.4 mini
      & 37.67 & 81.25 & 40.25 & 45.25 & 17.50 & 23.00 & 18.75
      & 58.46 & 85.75 & 66.75 & 36.25 & 39.75 & 50.25 & 72.00 \\
    Claude Sonnet 4.6
      & \underline{55.79} & \underline{93.00} & 64.50 & \textbf{50.25} & 36.00 & 46.00 & 45.00
      & \underline{66.38} & \textbf{93.75} & \underline{74.25} & \textbf{\underline{51.75}} & 48.00 & 50.00 & \textbf{80.50} \\
    Gemini 3.1 Pro
      & \textbf{56.88} & \textbf{94.50} & \underline{65.50} & \underline{48.25} & \textbf{\underline{53.75}} & 43.50 & 35.75
      & \textbf{\underline{69.46}} & \textbf{\underline{94.25}} & 70.75 & \textbf{49.00} & \textbf{\underline{63.75}} & \textbf{\underline{59.75}} & 79.25 \\
    Gemini 3.1 Flash
      & 40.17 & 91.75 & 40.00 & 42.75 & 23.00 & 20.00 & 23.50
      & 57.92 & 92.25 & 71.25 & 39.50 & 38.00 & 36.00 & 70.50 \\
    \midrule
    Average (Closed-Source)
      & 49.77 & 91.05 & 55.50 & 48.05 & 32.95 & 37.05 & 34.00
      & 63.76 & 91.90 & 72.30 & 44.40 & 49.55 & 47.35 & 77.05 \\
    \midrule
    \multicolumn{15}{c}{\textbf{Fine-tuned LLMs}} \\
    \midrule
    FT Llama-3.1-8B-Instruct
      & 53.13 & 92.75 & 63.50 & 39.75 & 36.00 & 41.50 & 45.25
      & 61.75 & 92.25 & 70.75 & 41.75 & 37.00 & \underline{54.25} & 74.50 \\
    FT Llama-3.1-8B-Instruct (error type)
      & 55.33 & 87.50 & \textbf{66.25} & 46.00 & 31.25 & \textbf{48.25} & \textbf{52.75}
      & 62.38 & 91.25 & 72.50 & \underline{46.00} & 33.50 & \textbf{55.75} & 75.25 \\
    \bottomrule[1.5pt]
    \end{tabular*}}
    \caption{Accuracy (\%) of LLMs on SDA-Synth under OEQ and MCQ. Task abbreviations: D=Descriptive, I=Inferential, E=Exploratory, P=Predictive, C=Causal, M=Mechanistic. \textbf{\underline{1st}}, \textbf{2nd}, \underline{3rd} highest scores per column are bold-underlined, bold, and underlined. FT=fine-tuned.}
    \label{tab:mcq-oeq-results}
  \end{table*}

\subsection{Validation and Dataset Statistics}
\label{sec:quality-control}

\paragraph{Validation.}
We apply the same validation pipeline to SDA-Synth and SDA-Real. We first check the consistency among each instance, dataset, derivation procedure, ground-truth answer, and task definition, removing 779 candidates with underspecified variables, dataset-instance mismatches, taxonomy errors, or ambiguous interpretations. We then conduct a manual quality audit of all 307 typed question templates, their associated programmatic derivation procedures, and 2,000 sampled instances. This process identified and corrected 11 minor issues and filtered out 324 unreliable cases. For multiple-choice instances, we recompute distractors with $f_{\tau}$ and remove 86 cases with non-unique correct answers or near-equivalent distractors.

\paragraph{Dataset statistics and splits.}
After validation, we obtain 6,231 SDA-Synth instances and 527 SDA-Real instances. For controlled evaluation on SDA-Synth, we stratify instances by five domains and six task types, keeping 200 instances for each domain-task pair and forming 6,000 instances. Each core instance has one MCQ version and one OEQ version. We split these instances into training (60\%) and test (40\%) sets while keeping paired formats together to prevent leakage. The training split is used only for fine-tuning, and all SDA-Real instances are reserved for independent real-data evaluation.

\section{Experiments}

\subsection{Experimental Setup}

Base-model evaluations on SDABench use a zero-shot setting. To make outputs comparable across models and response formats, we prompt each model to produce a structured JSON response containing two fields: a \texttt{reasoning} field for the step-by-step analysis and a \texttt{answer} field for the final result. Responses are parsed automatically; only the \texttt{answer} field is scored against the reference, while the \texttt{reasoning} field is used for error annotation (Section~\ref{sec:error-analysis}). The main results are reported on the held-out SDA-Synth test split, and we additionally evaluate models on SDA-Real under the same protocol.

\paragraph{Models.}
We evaluate 15 representative LLMs spanning closed-source and open-source models. \emph{Closed-source}: GPT-5.4~\cite{openai2026gpt54}, GPT-5.4 mini~\cite{openai2026gpt54mini}, Claude Sonnet 4.6~\cite{anthropic2026sonnet46}, Gemini 3.1 Pro~\cite{googledeepmind2026gemini31pro}, and Gemini 3.1 Flash~\cite{googledeepmind2026gemini31flashlite}. \emph{Open-source}: DeepSeek V3.2~\cite{liu2025deepseek}, DeepSeek R1~\cite{guo2025deepseek}, Qwen 3.5-397B-A17B~\cite{qwen3.5}, Qwen 3-235B-Instruct~\cite{yang2025qwen3}, Kimi K2.5~\cite{team2026kimi}, GLM 5~\cite{zeng2026glm}, Llama 3.3-70B-Instruct~\cite{grattafiori2024llama}, and Llama-3.1-8B-Instruct~\cite{grattafiori2024llama}. We additionally evaluate two LoRA fine-tuned variants of Llama-3.1-8B-Instruct trained on the SDA-Synth training split.

\paragraph{MCQ Evaluation.}
For multiple-choice questions, we use standard option accuracy: a response is correct only when the extracted option matches the reference label.

\paragraph{OEQ Evaluation.}
For open-ended questions, we score the normalized final answer using evaluators matched to the answer format: a numerical evaluator (relative-error tolerance), a textual evaluator (exact and structured-field matching), and an LLM evaluator (GPT-4o) as a fallback when deterministic matching fails.

\subsection{Main Results}
To evaluate the scientific data analysis capabilities of frontier LLMs, we analyze their performance across six scientific task types and five scientific domains.

\paragraph{Overall performance.}
Table~\ref{tab:mcq-oeq-results} reports overall results on the held-out SDA-Synth test split under both OEQ and MCQ formats. Because the four-option MCQ format has a 25\% chance floor, OEQ and MCQ accuracies are not directly comparable; we use OEQ as the primary metric and treat MCQ as a complementary signal.

Closed-source models set the strongest range, with GPT-5.4 at 58.33\%, Gemini 3.1 Pro at 56.88\%, and Claude Sonnet 4.6 at 55.79\% OEQ accuracy. Fine-tuned Llama-3.1-8B-Instruct approaches this range under both formats; the error-type objective further raises OEQ to 55.33\%, suggesting task-aligned supervision helps smaller models learn the benchmark's analysis structure.

Comparing the two formats reveals divergent behaviors. Kimi K2.5 on Predictive tasks achieves near-random MCQ accuracy (25.75\%) despite solving matched instances under OEQ, suggesting explicit options can induce local pattern matching. Conversely, Llama 3.3-70B-Instruct remains near the random baseline on several MCQ task types (causal 24.00\%, exploratory 23.50\%), indicating candidate options provide little benefit when the underlying analysis procedure is absent.

\noindent\textbf{Finding 1.}
  Frontier-scale models demonstrate the strongest analytical capabilities; however, open-ended scientific data analysis remains far from solved. Task-specific supervision can substantially narrow the gap between small open models and frontier closed-source models.

\paragraph{Performance across task types.}
Table~\ref{tab:mcq-oeq-results} shows a clear hierarchy across task types. Models perform best on Descriptive tasks, as they mainly require summarizing observed variables. Accuracy drops on Inferential and Causal tasks, which require selecting valid assumptions and reasoning over latent structure. The format gap is especially pronounced for Mechanistic tasks: MCQ options help strong models identify plausible mechanisms, whereas OEQ requires building the dependency structure by linking variables, relations, and intermediate causal steps. Exploratory and Predictive tasks remain difficult for different reasons: the former requires search over weakly constrained hypotheses, and the latter requires choosing an appropriate pattern and applying it to unseen cases.

\noindent\textbf{Finding 2.}
  LLMs are most effective at characterizing observed samples. Performance declines when tasks require selecting assumptions, modeling latent processes, or constructing mechanisms. A central difficulty is linking intermediate dependencies into a coherent chain of reasoning.

  \begin{table}[t]
    \centering
    {\small
    \setlength{\tabcolsep}{0pt}
    \renewcommand{\arraystretch}{1.08}
    \begin{tabular*}{\columnwidth}{@{\extracolsep{\fill}}lccccc@{}}
    \toprule
    \textbf{Task} & \textbf{Biology} & \textbf{Chemistry} & \textbf{Environment} & \textbf{Geography} & \textbf{Physics} \\
    \midrule
    Descriptive & 71.6 & 75.8 & 80.1 & 75.8 & 75.5 \\
    Inferential & 43.6 & 45.2 & 49.2 & 44.9 & 45.8 \\
    Exploratory & 35.6 & 37.4 & 40.7 & 30.9 & 33.8 \\
    Predictive & 22.5 & 27.6 & 34.9 & 29.6 & 29.5 \\
    Causal & 32.5 & 32.1 & 40.1 & 33.2 & 32.5 \\
    Mechanistic & 29.6 & 33.0 & 36.6 & 24.9 & 31.3 \\
    \midrule
    \textbf{Overall} & \textbf{39.2} & \textbf{41.9} & \textbf{46.9} & \textbf{39.9} & \textbf{41.4} \\
    \bottomrule
    \end{tabular*}}
    \caption{Average OEQ accuracy (\%) of LLMs on SDA-Synth across scientific data analysis tasks and scientific domains.}
    \label{tab:task-domain-oeq}
  \end{table}

\paragraph{Performance across scientific domains.}
Table~\ref{tab:task-domain-oeq} shows that domain-level variation reflects different reasoning burdens imposed by the underlying data structure, not merely terminology. Models achieve the highest overall OEQ accuracy in Environment (46.9\%), where many tasks involve smoother continuous variables and direct trend estimation. Biology shows the lowest overall accuracy, with a particularly low Predictive score (22.5\%), suggesting high-dimensional nonlinear interactions make out-of-sample generalization harder. Geography presents a different pattern: performance drops on Exploratory (30.9\%) and Mechanistic (24.9\%) tasks, requiring spatial reasoning from symbolic data.

\noindent\textbf{Finding 3.}
Domain variation reflects different reasoning burdens imposed by underlying data structure rather than mere terminology changes. Environment benefits from smoother continuous variables, while Biology suffers from high-dimensional nonlinear interactions. Geography struggles with spatial structure that must be inferred from symbolic data.

\begin{figure}[t]
  \centering
  \includegraphics[width=\linewidth]{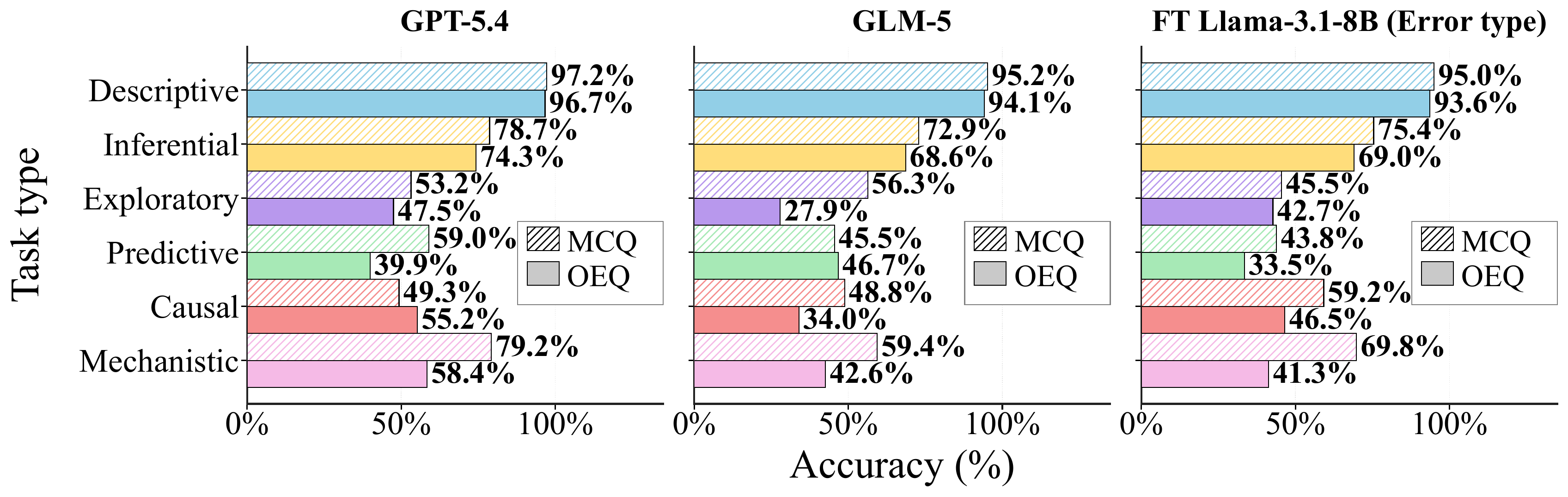}
  \caption{Accuracy (\%) of LLMs on SDA-Real across scientific data analysis tasks.}
  \label{fig:sda_real_mcq_qa}
\end{figure}

\paragraph{Results on SDA-Real.}
 Figure~\ref{fig:sda_real_mcq_qa} reports MCQ and OEQ accuracy for GPT-5.4, GLM 5, and fine-tuned Llama 3.1-8B-Instruct (Error type) across scientific data analysis tasks.
 SDA-Real broadly preserves the task-type hierarchy observed on SDA-Synth. Descriptive and Inferential tasks remain comparatively easier, whereas Exploratory and Predictive tasks remain more difficult. The full results are provided in the supplementary material C.

\begin{figure*}[t]
  \centering
  \includegraphics[width=\textwidth]{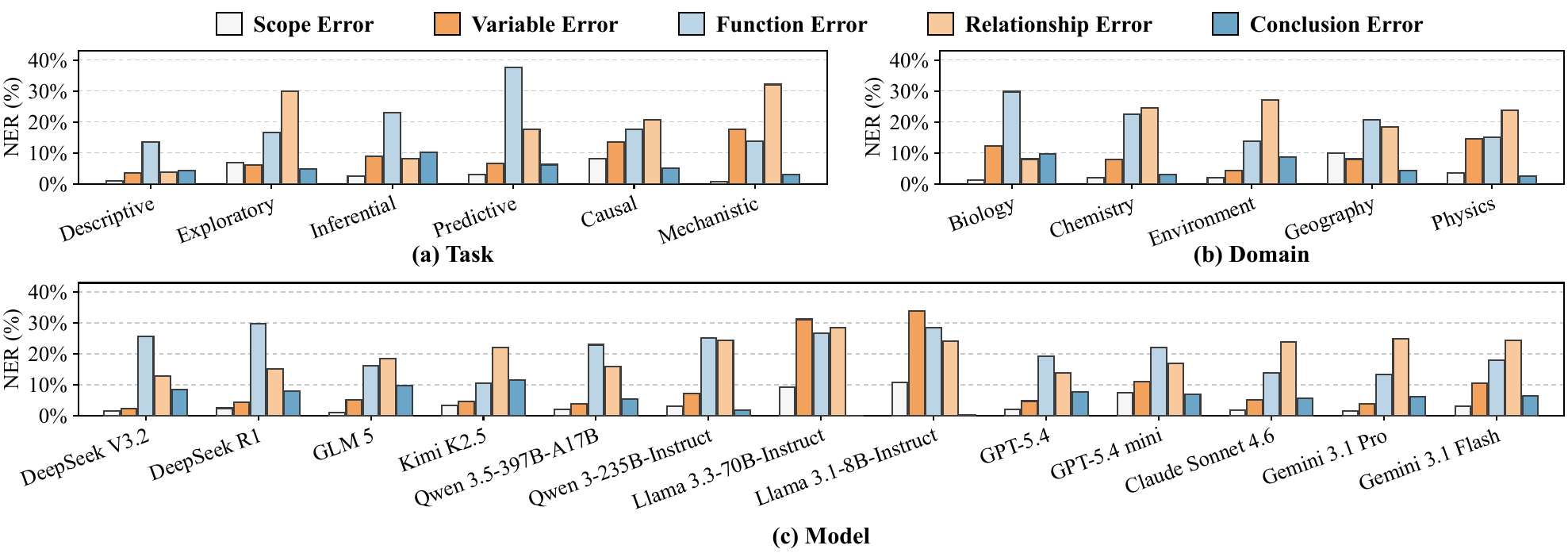}
  \caption{Normalized error rates across tasks, domains, and models on SDA-Synth.}
  \label{fig:error-analysis}
\end{figure*}

\subsection{Error Analysis}
\label{sec:error-analysis}
\paragraph{Error Analysis Framework.}
Aggregate accuracy cannot reveal where in the analysis execution process a model fails. We therefore conducted a bottom-up inspection of incorrect responses and identified five recurring failure modes ordered along the execution of scientific data analysis: (1) Scope Error, misinterpreting the analytical target or task scope; (2) Variable Error, selecting the wrong analysis variables; (3) Function Error, choosing or applying an incorrect analytical method; (4) Relationship Error, mischaracterizing the existence, direction, strength, or structure of variable interactions; and (5) Conclusion Error, drawing a final inference unsupported by the preceding analysis. We annotate the first error in the execution. To compare error profiles across tasks, domains, and models, we report the normalized error rate (NER), $\mathrm{NER}_{E}(S)=\tfrac{n_{E}(S)}{o_{E}(S)}$, where $S$ is a task, domain, or model subset; $n_E(S)$ is the number of errors labeled as type $E$; and $o_E(S)$ is the number of examples in which type $E$ could occur.

\paragraph{Errors across task types.}
Figure~\ref{fig:error-analysis}(a) shows that different task types exhibit distinct primary failure modes.
Function Errors peak in Predictive tasks (37.7\%) and remain high in Inferential (23.0\%) and Descriptive (13.4\%) tasks, reflecting inappropriate method selection or formula implementation. Relationship Errors dominate Mechanistic (32.1\%) and Exploratory (29.9\%) tasks, where the analytical target is the relational structure itself: models mistake noise for signal in Exploratory tasks and cannot reconstruct effect propagation without knowledge of the governing mechanism in Mechanistic tasks. Conclusion Errors are most prominent in Inferential tasks (10.2\%), reflecting misjudged significance under uncertainty.

\noindent\textbf{Finding 4.}
     Function Error dominates descriptive tasks, while Relationship Error becomes primary in more challenging task types, shifting the bottleneck from method selection to variable reasoning.

\paragraph{Errors across scientific domains.}
Figure~\ref{fig:error-analysis}(b) shows that LLM error patterns vary across scientific domains. In Biology, errors concentrate in Function Errors (29.8\%): models identify relevant entities but substitute generic statistical procedures for domain-specific analyses. Chemistry and Environment show Relationship Errors reaching 24.6\% and 27.0\%, respectively, as variable interactions depend on coupled processes, thresholds, or confounders models fail to capture. Geography shows the highest Scope Error rate (9.9\%), reflecting misinterpretations of the analytical target. Physics shows the highest Variable Error rate (14.5\%), where models retrieve correct equations but misalign them with variables, units, or physical roles.

\noindent\textbf{Finding 5.}
    Across domains, models fail not from lack of knowledge but from misapplication. They easily identify formulas and variables, yet overlook constraints like unit consistency, task boundaries, and variable interactions.

\paragraph{Errors across models.}

Figure~\ref{fig:error-analysis}(c) shows that models fail at different stages. Smaller baseline models (Llama 3.1-8B, Llama 3.3-70B) show the highest Variable Error rates (often above 30\%), failing to map questions to correct variables. Stronger models reduce these grounding errors, but their remaining failures move toward method execution: DeepSeek R1, DeepSeek V3.2, Qwen 3.5, and GPT-5.4 are dominated by Function Errors, while Kimi K2.5, Claude Sonnet 4.6, and Gemini 3.1 Pro have lower Function Error rates but higher Relationship Error rates, indicating they execute the intended operation yet still mischaracterize variable relationships.

\noindent\textbf{Finding 6.}
    As models improve, Scope and Variable errors decrease while Function, Relationship errors persist. This shifts error profiles toward later-stage failures, reflecting better early grounding rather than a decline in reasoning.

\section{Discussion}  

\paragraph{SDA-Synth versus SDA-Real.}
The broad preservation of the task-type hierarchy from SDA-Synth to SDA-Real suggests that synthetic semantic graphs capture the core reasoning demands of real scientific analysis. However, SDA-Real also exposes differences less visible in the controlled synthetic setting: real datasets use canonical variable names that align with pretrained knowledge (inflating mechanistic accuracy), while real exploratory analysis is more sensitive to heterogeneous schemas and noisy variables. These complementary strengths motivate keeping both components.

\paragraph{Limitations.}
SDA-Synth relies on synthetic semantic graphs whose mechanisms, while validated by domain experts, may not capture the full complexity of real experimental confounders and measurement artifacts. The MCQ format, despite perturbation-based distractors, can induce superficial pattern matching in some models. Our error taxonomy annotates the first error in the execution, which may undercount cascading failures.

\paragraph{Implications for future LLM development.}
Because later-stage errors persist even as grounding errors resolve with scale, future work should target three directions: assumption selection under uncertainty, relational reasoning over variable interactions, and iterative verification of intermediate results against task constraints. Beyond evaluation, SDABench's pipeline-ordered error taxonomy can guide adaptive training curricula that progressively target the capability gaps identified at each analysis stage.

\section{Conclusion}

We introduced SDABench, a capability-oriented benchmark that evaluates LLMs on scientific data analysis across six discovery capabilities and five domains. Evaluating 15 representative LLMs, we find that models handle descriptive analysis well but degrade sharply on tasks requiring assumption selection, latent-process modeling, or mechanistic reasoning. Stronger models do not lack scientific knowledge; rather, they apply it incorrectly. To localize \emph{where} models fail, we develop a five-stage error taxonomy (Scope, Variable, Function, Relationship, Conclusion) showing that Scope and Variable Errors decrease with model scale, whereas Function, Relationship, and Conclusion Errors persist, making later-stage reasoning failures the dominant bottleneck. This pipeline-ordered diagnosis enables targeted improvements beyond aggregate accuracy scores.

\bibliography{aaai2027}

\end{document}